\documentclass[conference]{IEEEtran}
\IEEEoverridecommandlockouts
\usepackage{tikz}
\usetikzlibrary{arrows.meta, positioning}
\usepackage{cite}
\usepackage{amsmath,amssymb,amsfonts}
\usepackage{algorithmic}
\usepackage{graphicx}
\usepackage{textcomp}
\usepackage{xcolor}
\usepackage{csquotes}
\def\BibTeX{{\rm B\kern-.05em{\sc i\kern-.025em b}\kern-.08em
    T\kern-.1667em\lower.7ex\hbox{E}\kern-.125emX}}

\newcommand\copyrighttext{%
  \footnotesize \textcopyright~\the\year{} IEEE. Personal use of this material is permitted.  Permission from IEEE must be obtained for all other uses, in any current or future media, including reprinting/republishing this material for advertising or promotional purposes, creating new collective works, for resale or redistribution to servers or lists, or reuse of any copyrighted component of this work in other works.%
}

\newcommand\copyrightnotice{%
\begin{tikzpicture}[remember picture,overlay]
\node[anchor=south,yshift=25] at (current page.south) {\begin{minipage}{0.9\textwidth}\centering\copyrighttext\end{minipage}};
\end{tikzpicture}%
}

\begin{document}

\title{Bayesian Optimization of Process Parameters of a Sensor-Based Sorting System using Gaussian Processes as Surrogate Models\\
}

\author{
    \IEEEauthorblockN{
        Felix Kronenwett,
        Georg Maier,
        Thomas Längle
    }
    \IEEEauthorblockA{
        \textit{Fraunhofer IOSB, Institute of Optronics, System Technologies and Image Exploitation, Karlsruhe, Germany} \\
         \{felix.kronenwett, georg.maier, thomas.laengle\}@iosb.fraunhofer.de}

}

\maketitle
\copyrightnotice
\begin{abstract}
Sensor-based sorting systems enable the physical separation of a material stream into two fractions. 
The sorting decision is based on the image data evaluation of the sensors used and is carried out using actuators. 
Various process parameters must be set depending on the properties of the material stream, the dimensioning of the system, and the required sorting accuracy.
However, continuous verification and re-adjustment are necessary due to changing requirements and material stream compositions. 
In this paper, we introduce an approach for optimizing, recurrently monitoring and adjusting the process parameters of a sensor-based sorting system. 
Based on Bayesian Optimization, Gaussian process regression models are used as surrogate models to achieve specific requirements for system behavior with the uncertainties contained therein.
This method minimizes the number of necessary experiments while simultaneously considering two possible optimization targets based on the requirements for both material output streams.
In addition, uncertainties are considered during determining sorting accuracies in the model calculation.
We evaluated the method with three example process parameters.%, object expansion in the spatial and temporal direction, as well as the delay time between sensor acquisition and actuators.
%The approach presented could significantly reduce the number of sorting experiments required to determine an optimum setting.
\end{abstract}

\begin{IEEEkeywords}
Sensor-Based Sorting, Process Optimization, Bayesian Optimization, Surrogate-Based Optimization, Gaussian Process Regression
\end{IEEEkeywords}

\section{Introduction}
Sensor-based sorting (SBS) systems are essential in various industries to efficiently separate material streams into different fractions based on sensor data evaluation and control of actuators. Typical fields of application include quality control in manufacturing processes, agricultural and food industries, mineral processing, and recycling operations. Examples of sorting tasks are the removal of fungus-infested kernels from seeds, the enrichment of ores in industrial minerals or the sorting of plastic waste according to polymer type. Hence, practical and precise sorting enhances product quality and significantly contributes to resource conservation and sustainability~\cite{maier2024survey}.

To achieve optimal sorting results, it is necessary to parameterize process parameters carefully~\cite{gulcan2020novel}. Optimizing these parameters presents a complex challenge due to the intricate interactions between the sensors and actuators and the variability in material streams. The key parameters for the configuration of the actuators for ejection must be carefully selected depending on the material properties, the system dimensions, and the desired sorting accuracy, and also adjusted during sorting operation if the properties of the material stream change. Manual adjustments by experts are time-consuming and may not effectively adapt to changes, such as variations in loading density or material properties, such as size, weight, or shape.

\begin{figure}[t]
\centerline{\includegraphics[width=\columnwidth]{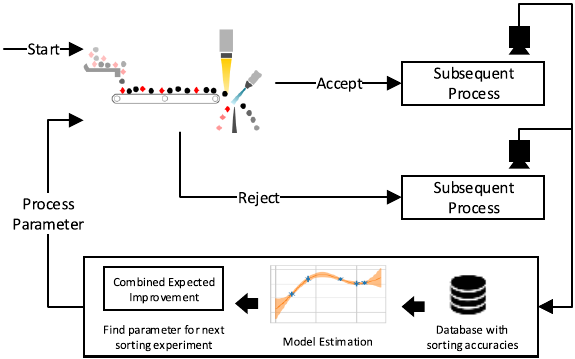}}
\caption{Overview of process parameter exploration of a sensor-based sorting (SBS) system using Bayesian optimization (BO). Starting from an initial set of parameters, the resulting accept and reject material streams are analyzed and new parameters for subsequent sorting experiments are proposed using Gaussian process regression (GPR) models.}
\label{fig/Process_parameter_opt_SBS}
\end{figure}

This paper introduces a methodology that employs Bayesian optimization (BO) using Gaussian process regression (GPR) models as surrogate models to optimize and recurrently adjust the process parameters of SBS systems. An overview of the approach is provided in Figure \ref{fig/Process_parameter_opt_SBS}. By capturing the complex interactions between the process parameters and sorting results, this approach allows for efficient exploration of the parameter space and improves sorting performance, even under uncertainties and variable conditions. This method accounts for the inherent noise in the observed data, such as measurement variances due to material throughput fluctuations and non-deterministic ejection behaviors caused by object variations and system jitter. Our approach makes it possible to consider the requirements for both material fractions resulting from the sorting process during optimization, monitor them continuously, and adjust them in-line if necessary.

\subsection{Related Work}
Optimization problems for adapting process parameters in engineering often have a high dimensionality and a nonlinear, non-convex behavior and are, therefore, quite complex. Surrogate-based optimization (SBO) is a method for reducing numerical effort during parameter optimization~\cite{forrester2008engineering}. Surrogates are approximations of the behavior of complex systems and are employed to reduce computation time and resources. In existing work, BO is often used with a Gaussian process (GP) as a surrogate model. BO is a strategy for the global optimization of an unknown objective function, which is expensive to evaluate~\cite{jones1998efficient, snoek2012practical}. GPR is used to model the process and offers the possibility of providing uncertainty in predicting the process parameter values. Practical applications for process parameter optimization that use BO include the search for parameters in the synthesis of polymer fibers~\cite{li2017rapid}, parameter optimization of powder film drying processes~\cite{nagai2022sample}, and optimization of a grinding machine process, considering quality, cost, and safety constraints~\cite{maier2020self}. Further work in the field of process parameter optimization uses neural networks as surrogate models~\cite{PFROMMER2018426}. Furthermore, research is being conducted on process parameter optimization employing reinforcement learning~\cite{ZIMMERLING2022110423}. These methods have particular strengths in the modeling of highly complex processes in variable situations.

Several studies have investigated the effects of different process parameters on the sorting process. 
In several cases, simulation models have been used for this purpose.
For example, in~\cite{pascoe_prediction_2015}, a Monte Carlo simulation was used to determine the resulting sorting quality for different mass flows.
The results suggest that sorting quality decreases exponentially as mass flow increases.
However, the authors did not consider the varying mass flows over time.
In~\cite{bauer2022towards}, a multitude of potentially varying process parameters, such as mass flow and particle velocity, were considered to determine the optimal operation points of a sorting system. 
The authors employed a coupled discrete element method and computational fluid dynamics to numerically simulate the sorting process. 
The same type of simulation was used in~\cite{vieth2023improving}.
Here, the authors modeled a sorting system with recirculation and introduced a closed-loop controller to improve the sorting quality.
This work was further extended in~\cite{walker2025stochastic}, where the authors proposed a stochastic model of the sorting system in combination with a stochastic nonlinear model predictive controller for closed-loop control of the sorting system.

\section{Sensor-Based Sorting Systems}
\subsection{Functional Principle}

SBS systems offer solutions for physically separating a material stream into predefined classes. In many cases, the aim is to remove inferior parts from a material stream, resulting in a binary sorting task. The task can therefore be understood as an \textit{accept-or-reject} decision.
Material is fed onto a transportation mechanism, for instance, a conveyor belt or chute, and observed by one or multiple sensors during transportation at the so-called inspection line. 
While the choice of sensors depends on the sorting task at hand, i.e., which material properties are to be made \enquote{visible}, line-scanning, imaging sensors are commonly used. 
The captured image data are processed to localize and classify individual objects.
The classification results serve as the basis for the sorting decisions. 
Typically, a bounding box represents the size and position of each object. 
Fast-switching pneumatic valves are generally used to separate individual objects from the material stream.
The valves are arranged in a row orthogonal to the main transport direction, e.g., the running direction of the conveyor belt. 
Individual nozzles can be activated to eject individual objects. 

Assuming a constant speed for the bulk material during transport on the conveyor belt, a fixed delay time between sensor detection and ejection can be set. 
Both the number of activated nozzles and the duration of activation are adjustable and depend on the size of the detected objects.

The sorting quality achieved depends primarily on the classification accuracy of the bulk material objects and correct physical ejection. 
Depending on the material and structure, various process parameters can influence and optimize the timing and accuracy of the separation. 

\subsection{Process Parameters in Sensor-Based Sorting Systems}
An SBS system consists of many process parameters that influence the sorting process and, therefore, have a significant impact on the quality of the sorting result, both in terms of the reject and the accept fraction~\cite{gulcan2020novel}. 
The process parameters are configured primarily during the system commissioning.
Adjustments are made if it is observed during the operation that the setting no longer delivers the expected results.
Changes in sorting behavior can be caused by changes in the input material flow, such as alterations in the loading density, or varying material properties, such as size, weight, or shape.

The results of the sensor data evaluation are the identification and localization of the individual bulk material objects in the form of a class label and bounding box, defined as $(x_{\text{min}}, y_{\text{min}}, x_{\text{max}}, y_{\text{max}})$, where $x_{\text{min}}$ and $x_{\text{max}}$ represent the horizontal limits of the object on the conveyor belt, and $y_{\text{min}}$ and $y_{\text{max}}$ represent the vertical limits of the object as detected by the imaging sensor (example in~\cite{kronenwett2024sensor}).

Our investigations focus on three relevant process parameters that significantly influence sorting behavior and presumably have the most substantial impact.
All parameters are related to the interaction between the analyzed imaging sensor data and actuators used to eject defined material classes or objects using a pneumatic nozzle array:
\begin{itemize}
    \item \textbf{Reaction Lines ($T_R$)}: The delay time $T_R$ between the first appearance of an object under the inspection line of a line-scanning imaging sensor and the extraction actuator. The delay when opening or activating the compressed air nozzle, represented by $\Delta T_{\text{nozzle}}$, and the general jitter $\Delta T_{\text{jitter}}$ should not be neglected when setting the parameters. The effective reaction time can be expressed as $T_{\text{effective}} = T_R + \Delta T_{\text{nozzle}} + \Delta T_{\text{jitter}}$.
    \item \textbf{Extended Time ($T_E$)}: The parameter $T_E$ denotes the enlargement of the bounding box of a detected object along the time axis, perpendicular to the inspection line. The result is an extension of the nozzle activation during ejection. This influence can be described as the temporal extension $\Delta T_{E}$ applied to the initial bounding box time dimension, $S_{\text{box}}$. Thus, the extended time is $T_{\text{extended}} = T_{\text{box}} + \Delta T_E$.
    \item \textbf{Extended Space ($S_E$)}: The parameter $S_E$ indicates the enlargement of the bounding box of a detected object vertically to the conveyor belt movement. This results in a broadened ejection window, potentially activating more compressed air nozzles during the ejection. The spatial extension $\Delta S_{E}$ is applied to the initial bounding box spatial dimension $S_{\text{box}}$, leading to $S_{\text{extended}} = S_{\text{box}} + \Delta S_E$.
\end{itemize}
Thus, the mathematically redefined bounding box considering $T_E$ and $S_E$ extensions can be expressed as
\begin{equation*}
    [x_{\text{min}} - \frac{\Delta S_E}{2},\ y_{\text{min}} - \frac{\Delta T_E}{2},\ x_{\text{max}} + \frac{\Delta S_E}{2},\ y_{\text{max}} + \frac{\Delta T_E}{2}] \text{.}
\end{equation*}
The line-scan camera delivers image lines in a fixed cycle.
This line clock serves as the time coordinate in the proposed system.
Hence, the units for $T_{R}$ and $T_E$ are given in lines ($L_{T_{R/E}}$), which can be converted to time with the recording frequency, using $T_{R,\text{time}} = \frac{L_{T_r}}{f_{\text{line}}}$ and $T_{E,\text{time}} = \frac{L_{T_e}}{f_{\text{line}}}$.
The accuracy of the parameter $S_E$ usually is limited to the length of one pixel.

When manually adjusting the three parameters, a compromise must be made between the desired sorting objectives.
If the goal is to eject and accurately target as many \textit{reject} objects as possible, high values for the parameters $T_E$ and especially $S_E$ are suitable; however, with high material occupancy, more acceptable parts are also ejected.
Therefore, the objectives of the two resulting \textit{reject} and \textit{accept} material streams must be defined based on the economic requirements and demands of subsequent processing operations.

\subsection{Sorting Accuracy Determination}
To optimize the process parameters, a feedback loop involving the current sorting results and the desired sorting accuracies is required.
We assume an SBS system is integrated into a processing chain with subsequent systems and processes (see Figure \ref{fig/Process_parameter_opt_SBS}).
By leveraging the sensor data from this setup, we can determine the sorting quality effectively.

Two imaging sensors are utilized: one observes the discharged material stream (accepted objects), and the other monitors the unaffected material stream (rejected objects). 
These sensors detect bulk material objects after sorting, so the sorting quality can be evaluated using a confusion matrix. The process is as follows:
\begin{enumerate}
    \item Using imaging sensors to monitor the two resulting material streams after sorting.
    \item Analyzing the image data and counting objects based on their classes to determine their distribution in each stream.
    \item Compiling the results into a confusion matrix with four categories:
    \begin{itemize}
        \item True Positives (TP): Number of objects that belong to the accept class and were correctly sorted into the accept stream.
        \item False Negatives (FN): Number of objects that belong to the accept class but were incorrectly sorted into the reject stream.
        \item True Negatives (TN): Number of objects that belong to the reject class and were correctly sorted into the reject stream.
        \item False Positives (FP): Number of objects that belong to the reject class but were incorrectly sorted into the accept stream.
    \end{itemize}
\end{enumerate}
The resulting confusion matrix is expressed as:
\begin{equation*}
    \begin{array}{c|cc}
              & \text{Predicted Accept} & \text{Predicted Reject} \\
    \hline
    \text{Actual Accept} & \text{TP} & \text{FN} \\
    \text{Actual Reject} & \text{FP} & \text{TN} \\
  \end{array}
\end{equation*}
The total number of objects $N_{\text{total}}$ is calculated as the sum of all detected objects.
The accuracy of the sorting process is then determined according to
\begin{equation}
    \text{Accuracy} = \frac{\text{TP} + \text{TN}}{N_{\text{total}}} \text{.}
\end{equation}
This metric provides a quantitative assessment of the  sorting system's performance, enabling targeted adjustments to improve overall accuracy.
To evaluate the accept and reject streams separately, the normalized TP and TN values can be used, which are calculated by dividing the number of objects classified by the sensors into the accept or reject classes:
\begin{equation}
    \text{TP}_n = \frac{\text{TP}}{\text{TP} + \text{FN}} \text{, } \text{TN}_n = \frac{\text{TN}}{\text{TN} + \text{FP}}\text{.}
\end{equation}

Assuming constant sorting rates and processing rates, as well as the independence of the objects' processing, a correlation between variance and duration of an observation interval or a measurement can be determined. Since we count the number of objects over a specific period, the number of TP and TN follows a binomial distribution.
The total number of observed objects $n$ was proportional to the data collection time $t$ (assuming a constant processing rate $r$).
It turns out that the variance of the TP and TN measurements are inversely proportional to the data acquisition time $t$ as given by
\begin{equation}
    \text{Var}(\hat{p}) \propto \frac{1}{t} \text{.}
    \label{eq:var}
\end{equation}
The longer the data collection time, the greater the number of observed objects $n$, and the smaller the variance of the estimated TP rate.
Therefore, a longer data collection time leads to more reliable and stable TP rate measurement.
This is crucial for evaluating the sorting system's performance because a lower variance allows for better differentiation between random fluctuations and actual system changes.

In both the actual data evaluation of the imaging sensors for the actual sorting process and the downstream data evaluation to determine the sorting quality, it is assumed that there is no sensory false identification of the objects.
While this is obviously unrealistic and not the case, we argue that depending on the material flow, false classification rates are very low in relation to the material or object throughput, and their influence on the calculated sorting quality is, therefore, negligible.

The variance of the accuracy due to misclassifications can be approximated by
\begin{equation}
    \text{Var}(\text{Accuracy}) = \frac{p(1-p)}{N_{\text{total}}} \text{,}
\end{equation}
where $p$ is the true classification accuracy rate.

In cases with very few misclassifications and a large $N_{\text{total}}$, the variance becomes negligible, indicating minimal impact on the measured sorting accuracy.
Therefore, with many observations and few errors, sorting accuracy remains a reliable metric for system performance evaluation.

\subsection{Data Acquisition}
An in-house lab-scale sorting system with results and physical properties that can be directly scaled up to larger systems was utilized in our study.
Bricks and sand-lime bricks were selected as exemplary bulk material streams.
Because of the different color characteristics (red and white), the misclassifications in the image data evaluation can be assumed to be close to zero.
The two material streams, i.e., accepted and rejected, were also recorded after sorting using the same camera model and evaluation algorithm to describe the sorting accuracy using the confusion matrix described above.
The observation period for parameter configuration was divided into several parts to monitor the variance of the results over time.

The following values were determined for the three system parameters and collected over a longer period:
\begin{itemize}
    \item $T_R = \{12,\ 13,\ 14,\ 15,\ 16,\ 17,\ 18,\ 19,\ 20,\ 21\}$
    \item $T_E = \{0,\ 2,\ 4,\ 6,\ 8\}$
    \item $S_E = \{0,\ 2,\ 4,\ 6,\ 8\}$
\end{itemize}
The data units are camera lines or pixels; however, for readability, they are neglected in the remainder.
The total set of parameter combinations is the Cartesian Product of these sets, $T_R \times T_E \times S_E$, resulting in 250 possible combinations and sorting experiments.

The measurement data were collected for approximately 3 to 5 minutes for each parameter configuration.
The number of objects was determined every 10 seconds. For this time span a confusion matrix was created, and the derived metrics were calculated.
This setting is a compromise depending on the set material throughput, the reduction of the measurement noise according to (\ref{eq:var}), and the competing reduction of the optimization process duration.
The final sorting performance is determined by averaging over the entire measurement interval.
The noise in the measured values can be estimated using shorter statistical intervals.

\subsection{First Examination of the Data}
\begin{figure}[tbp]
\centerline{\includegraphics[width=\columnwidth]{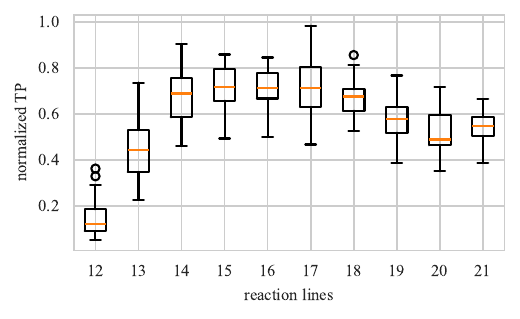}}
\centerline{\includegraphics[width=\columnwidth]{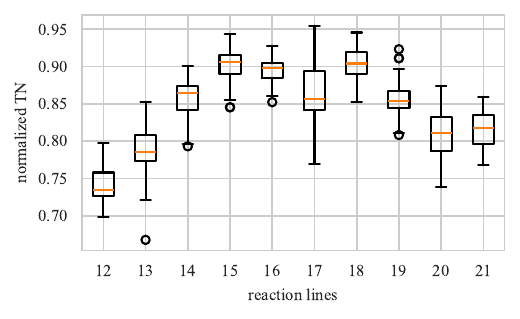}}
\caption{Box plot of the normalized TP (top) and TN (bottom), with an acquisition interval of 10 seconds during the sorting process as a function of the reaction lines parameter with constant other process parameter values (expanded time and space are set to zero).}
\label{fig/box_plot}
\end{figure}

Figure \ref{fig/box_plot} shows the normalized TP and TN metrics as a function of the reaction lines parameter (other parameter values were constantly set to zero) of the acquired data as a box plot.
The variance of the normalized TP measurements was significantly lower because of the higher number of objects in the accepted material flow, which led to more accurate measurements.
The reject material flow is significantly lower, so the physical ejection errors have a greater direct influence on the metric.

When looking at the reaction lines parameter in isolation, it can be seen that the normalized TP value resembles a quadratic function with a clear maximum.
The timing for the best ejection point is good only at one point; if ejected too early or too late, no object can be hit efficiently.
However, the relationship with the other two parameters is less easy to explain and visualize.
For this reason, a GPR model is used, which can better recognize the relationships.

\section{Methodology}
    
\subsection{Problem Definition}
The optimization of the process parameters of SBS systems differs from other problems in process parameter optimization.
One challenge is the noise observed in the experiment, i.e., the sorting accuracy measured in the subsequent processes.
The calculated accuracy metrics were subject to substantial variations, mainly depending on the material throughput within the measurement interval and class distribution.
At the same time, the ejection behavior is not deterministic due to individual object sizes, shapes, flight behavior, jitter in the control of the compressed air nozzles, and many other factors.
After a sorting experiment with defined process parameters, the result is subject to uncertainties, different from other studies and process parameter optimization examples.

Another special feature in the context of SBS systems is that the two output streams, accepted and rejected, can be subject to different requirements.
For example, the accept stream should be as pure as possible and free from foreign objects, whereas the composition of the reject stream is irrelevant.
However, other requirements are possible, depending on the individual sorting requirements.
Therefore, optimization must make it possible to weigh the requirements for the output streams individually.

\subsection{Bayesian Optimization using a Gaussian Process as Surrogate Model}
BO is a powerful strategy for the global optimization of black-box functions, which are expensive to evaluate and lack analytical expressions.
It is beneficial in scenarios where function evaluations are costly, such as hyperparameter tuning in machine learning models, engineering design optimization, and experimental sciences~\cite{shahriari2015taking}.

At the core of the BO, there are two primary components: a surrogate probabilistic model and an acquisition function.
The surrogate model, often a GP, approximates an unknown objective function based on prior observations.
The acquisition function leverages the predictive distribution of the surrogate model to determine the next point to be evaluated by balancing exploration and exploitation.

A GP is a distribution over functions, such that any finite set of function values $f (x_1), f (x_2), . . . f (x_N)$ have a joint Gaussian distribution.
It is fully specified by its mean function $m(\mathbf{x})$ and covariance function $k(\mathbf{x}, \mathbf{x}')$ according to
\begin{equation}
    f(\mathbf{x}) \sim \mathcal{GP}(m(\mathbf{x}), k(\mathbf{x}, \mathbf{x}')) \text{.}
\end{equation}
The covariance function also called the \textit{kernel}, defines the relationship between function values at different points according to
\begin{align}
    Cov[f(x),f(x')] &= k(\mathbf{x}, \mathbf{x}') \\
    &= \mathbb{E}[(f(\mathbf{x}) - m(\mathbf{x}))(f(\mathbf{x}') - m(\mathbf{x}'))] \text{.}
\end{align}
Typically, the mean function is assumed to be zero: $m(\mathbf{x}) = \mathbb{E}[f(\mathbf{x})] = 0$. Uncertainty regarding the mean function can be considered by adding an extra term to the kernel~\cite{10.7551/mitpress/3206.001.0001}.

Given a set of observations $\mathcal{D} = \{ (\mathbf{x}_i, y_i) \}_{i=1}^n$, where $y_i = f(\mathbf{x}_i) + \epsilon$ and $\epsilon \sim \mathcal{N}(0, \sigma_n^2)$ represents noise, the GPR model can be updated to provide predictions at new points with associated uncertainties~\cite{brochu2010tutorial}.
When a GP is applied to regression problems with observed data, this approach is referred to as GPR.

The expected improvement (EI) is a common acquisition function that balances the potential improvement over the current best observation, as given by
\begin{align}
    \text{EI}(\mathbf{x}) = &(\mu(\mathbf{x})-f_{\text{best}}) \cdot \nonumber \Phi\left( \frac{\mu(\mathbf{x})-f_{\text{best}}}{\sigma(\mathbf{x})} \right)\\
    &+ \sigma(\mathbf{x}) \phi\left( \frac{\mu(\mathbf{x})-f_{\text{best}}}{\sigma(\mathbf{x})} \right) \text{,}
\end{align}
where $\mu(\mathbf{x})$ and $\sigma(\mathbf{x})$ are the predictive mean and standard deviation at $\mathbf{x}$, $f_{\text{best}}$ is the best-observed value, and $\Phi(\cdot)$ and $\phi(\cdot)$ are the cumulative and probability density functions of the standard normal distribution, respectively.

\subsection{Weighted Multi-Objective Bayesian Optimization with Two Gaussian Processes and a Combined Expected Improvement}
In the following section, the optimization process for the process parameters of an SBS system is introduced, in which different requirements can be placed on the two output material streams.
An efficient strategy for simultaneously optimizing the two requirements, which may work against each other, is to calculate two separate GPR models for each output: the accept and reject streams.

The requirements for the accept and reject material streams are combined by calculating the introduced combined EI, in which the objectives for the accuracy of both material streams are weighted.
Therefore, the EI must first be determined for each separate GPR model:
\begin{equation}
\mathrm{EI}_{\text{comb}}(x)
= w_{\text{a}}\,\mathrm{EI}_{\text{accept}}(x)
+ w_{\text{r}}\,\mathrm{EI}_{\text{reject}}(x)
\label{eq:ei_comb}
\end{equation}
with weights $w_{\text{accept}},w_{\text{reject}} \ge 0,\ w_{\text{accept}} + w_{\text{reject}} = 1$.
The parameter values for the highest combined EI were used as the settings for the subsequent sorting experiment:
\begin{equation}
X_{\text{next}} \;=\; \arg\max_{x\in S}\ \mathrm{EI}_{\text{comb}}(x).
\label{eq:xnext}
\end{equation}
Furthermore, an adjustment is made to consider the uncertainty of the results of the sorting experiments in the GPR model calculations.
We treat each measurement’s variance as observation noise. During GP training, we incorporate it by adding the scaled per-sample variances ($\lambda\in[0,1]$) to the kernel diagonal:
\begin{equation}
K_{\text{neu}} \;=\; K(X,X)\;+\;\mathrm{diag}\!\big(\lambda\,\sigma_1^{2},\ldots,\lambda\,\sigma_n^{2}\big).
\label{eq:hetero_lambda}
\end{equation}
The factor $\lambda$ controls how strongly the empirical noise estimates influence the fit.

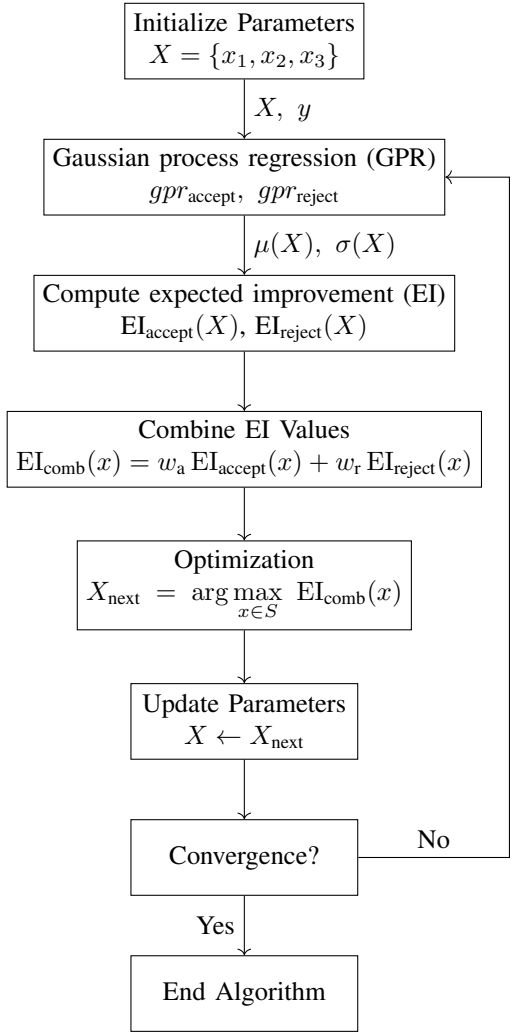
\begin{figure}[htbp]
    \centering
    \begin{tikzpicture}[
        node distance=1.8cm, 
        auto, 
        align=center,
        box/.style={rectangle, draw, minimum width=3cm, minimum height=1cm}
        ]
      % Nodes
      \node [box] (start) {Initialize Parameters\\ $X = \{x_1, x_2, x_3\}$};
      
      \node [box, below of=start] (gpr) {Gaussian process regression (GPR)\\
      $gpr_{\text{accept}},\ gpr_{\text{reject}}$};
      
      \node [box, below of=gpr] (ei) {Compute expected improvement (EI)\\
      $\text{EI}_{\text{accept}}(X)$, $\text{EI}_{\text{reject}}(X)$};
      
      \node [box, below of=ei] (combine) {Combine EI Values\\
      $\mathrm{EI}_{\text{comb}}(x)
= w_{\text{a}}\,\mathrm{EI}_{\text{accept}}(x)
+ w_{\text{r}}\,\mathrm{EI}_{\text{reject}}(x)$};
      
      \node [box, below of=combine] (optimize) {Optimization\\
      $\displaystyle X_{\text{next}} \;=\; \arg\max_{x\in S}\ \mathrm{EI}_{\text{comb}}(x)$};
      
      \node [box, below of=optimize] (update) {Update Parameters\\ $X \leftarrow X_{\text{next}}$};
      
      \node [box, below of=update] (check) {Convergence?};
      
      \node [box, below of=check] (exit) {End Algorithm};
      
      % Arrows
      \draw [->] (start) -- node[right] {$X,\ y$} (gpr);
      \draw [->] (gpr) -- node[right] {$\mu(X),\ \sigma(X)$} (ei);
      \draw [->] (ei) -- (combine);
      \draw [->] (combine) -- (optimize);
      \draw [->] (optimize) -- (update);
      \draw [->] (update) -- (check);
      \draw [->] (check.east) -- node[midway, above] {No} ++(2cm,0) |- (gpr);
      \draw [->] (check) -- node[midway, left] {Yes} (exit);
      
    \end{tikzpicture}
\caption{Sequence of the Bayesian optimization (BO) approach of the process parameters of a sensor-based sorting system using two separate Gaussian process regression (GPR) models for the accepted and rejected material streams and a combined expected improvement (EI).}
\label{ablauf}
\end{figure}

The final sequence for optimizing the process parameters of an SBS system is shown in Figure \ref{ablauf}, where $X = (x_1, x_2, x_3)$ represents the parameter sets for \textit{reaction lines}, \textit{extended time}, and \textit{extended space}.
The optimization begins with a set of several initial process parameter configurations and experiments. The results of these experiments were then used to calculate the initial GPR models.

\section{Application of the Optimization Strategy to the Process Parameters of a Sensor-Based Sorting System}
The optimization process was evaluated using the presented dataset and sorting scenario.
After a defined set of initial experiments, two GPR models were calculated using the obtained normalized TP and TN metrics, as well as the estimation of the measurements' variance.
The combined EI was calculated using the separate EI of the two models, which were scaled with a weight of $0.5$ for this test.
Subsequently, using the newly derived parameter configuration, a new experiment can be conducted in the form of a time-limited sorting process.
The database containing the sorting results was then expanded.

\begin{figure*}[htbp]
\centerline{\includegraphics[width=\textwidth]{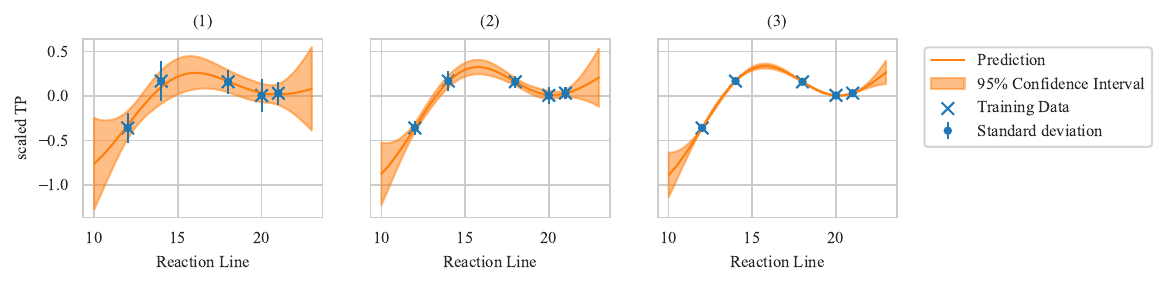}}
\caption{Results of the Gaussian process regression (GPR) models for predicting the normalized and scaled TP value based on the reaction lines parameter for different weightings of the variance in the calculation of the GPR models. (1): $\lambda = 0.1$, (2): $\lambda = 0.01$ and (3): $\lambda = 0$.}
\label{fig/rl_std}
\end{figure*}

Uncertainty is considered in the kernel in the form of the variance of the individual measurement intervals during the sorting process to calculate the GPR models.
The influence on the kernel must be weighted due to dependencies on the duration of the acquisition intervals and the material throughput rate.
Using the example of the single isolated parameter reaction lines ($T_R$), the variance of the normalized TP metric and the effects on the GPR are plotted in Figure \ref{fig/rl_std} for different scalings using the weight $\lambda$ and a small subset of experiments.
If the weighting in the kernel is too high ($\lambda \geq  1$), the result is generalized with high uncertainties overall.
Values far less than one ($0 \leq \lambda \ll 1$) or even zero enable a better-fitting model despite the relatively shorter acquisition interval of the sorting result.
However, overfitting is highly likely to occur.
Therefore, the variance estimation for the GP model calculations was weighted with $\lambda = 0.1$, offering the best generalization.

The boundaries of the search space can be freely determined when calculating the EI value for both models.
It is recommended and generally necessary to include prior knowledge of which parameter intervals are reasonable.
For the evaluation, the search space was constrained to the difference between the limit values of the initial start parameter set plus minus one.

The initial number of experiments was set to 12, comprising a combination of three reaction lines parameters and two extended space and time parameters:
\begin{equation*}
    [T_R, T_E, S_E]_{\text{start}} = \left[ \{12, 18, 21\}, \{0, 8\}, \{0, 8\} \right]\text{.}
\end{equation*}
The number of parameters determines the duration of the initial experiments.
Therefore, they should be as few as possible and cover the parameter space as well as possible.
Expert knowledge can be used in the selection process to reduce the number of options and minimize the exploration space.
Table \ref{tab1} lists the proposed parameters after calculating the combined EI after the initial sorting experiments using $[T_R, T_E, S_E]_{\text{start}}$ and each subsequent experiment.
The settings in the sorting system software only allow integer values.
Therefore, parameter values must be rounded to existing adjustable values if necessary.
After three steps, the optimization converges, yielding the final parameter set.
It can be seen that the reaction lines parameter converges purposefully, whereas the relationships between the extended time and extended space parameters appear more challenging.
\begin{table}[tbp]
\caption{Resulting next process parameter set $X$ for each optimization step after maximizing the combined expected improvement (EI) using all combinations of the starting parameter set $[T_R, T_E, S_E]_{start} = [\{12, 18, 21\}, \{0, 8\}, \{0, 8\}]$.}
\begin{center}
\begin{tabular}{|c|c|c|c|}
\hline
\textbf{Step}&\multicolumn{3}{|c|}{\textbf{$\arg\max_X\ \text{combined\ EI}(X)$}} \\
\cline{2-4} 
\textbf{Number} & \textbf{\textit{Reaction Lines}}& \textbf{\textit{Extended Time}}& \textbf{\textit{Extended Space}} \\
\hline
1 & 15.67 & 7.51 & 2.29 \\
\hline
2 & 14.57 & 0.82 & 6.04 \\
\hline
3 & 14.76 & 1.80 & 6.20 \\
\hline
\end{tabular}
\label{tab1}
\end{center}
\end{table}

Using all measured values from the dataset, a GPR model was calculated to derive the optimum parameter configuration and thus to assess the result of the recursive optimization.
The optimal result calculated from the maximum value of the calculated regression model is $[T_R, T_E, S_E]_{best} = [15.27,  1.29,  6.30]$.
The optimum configuration of the reaction lines parameter can already be read off well from the collected measured values and appears generally and mechanically independent of the other parameters.
The most robust value is the time point of the object center point of the compressed air nozzle (neglected by the jitter of the nozzle activation).
The extended time was short, and the objects were similar in shape and approximately complex.
Due to possible transverse movements between the sensor line and nozzle array, it is recommended to select a larger extended space to reliably hit and eject objects.

\begin{figure*}[htbp]
\centerline{\includegraphics[width=\textwidth]{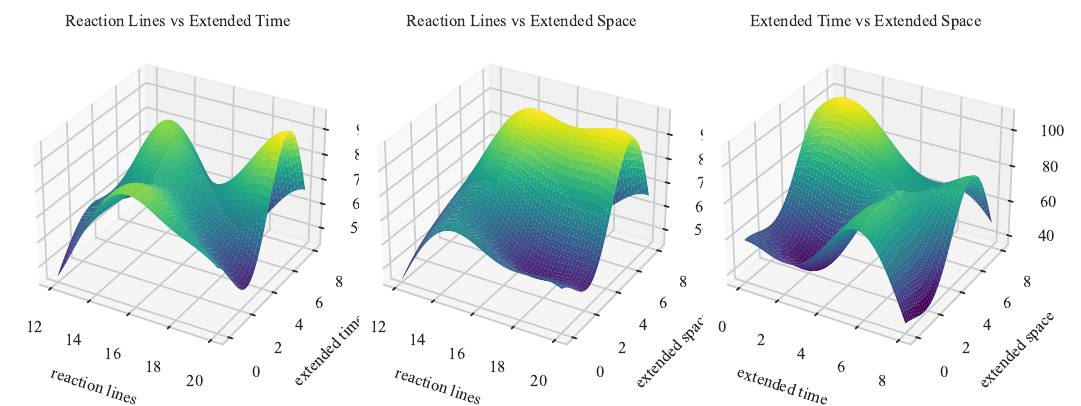}}
\centerline{\includegraphics[width=\textwidth]{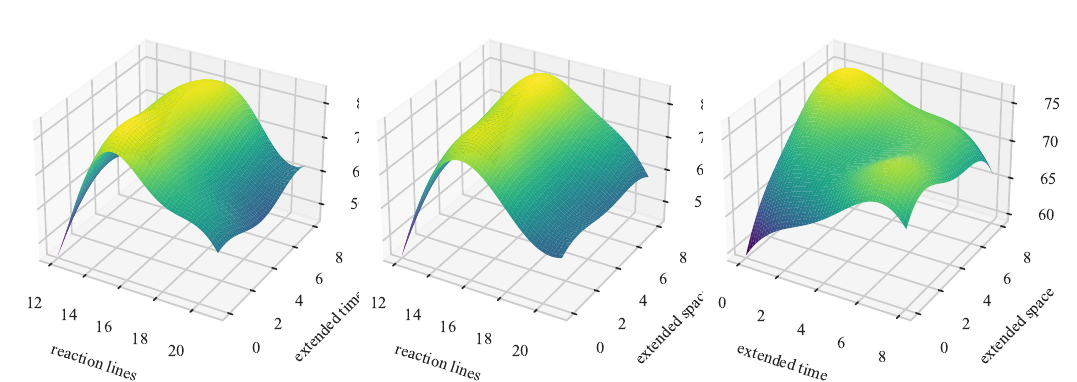}}
\caption{Evolution of the Gaussian process regression (GPR) model for the normalized TP value. The mean of the predicted TP values is plotted against two of the three process parameters for each case. 1st row: GPR model after the initial sorting experiments; 2nd row: GPR model with all TP values from the dataset.}
\label{fig/GPR_evolution}
\end{figure*}

The evolution of one of the GPR models, namely the model of the normalized TP value, is shown in Figure \ref{fig/GPR_evolution}.
For the first GPR model, which was calculated using the values from the initial sorting experiments, two of the three process parameters were plotted against the mean of the predicted TP values.
The second diagram illustrates the GPR model derived using the complete dataset encompassing all measurements.
Although the GPR model is initially imprecise and appears to have several local maxima and minima, the ideal model, which was iteratively adjusted at each optimization step, is smoother.

For further investigation, the optimization strategy was applied once more with the same starting set of process parameters, but with different prioritization of the SBS process's output stream properties.
This behavior was achieved by selecting distinct weight values when calculating the combined EI.
The results are presented in Table \ref{tab2}.
A higher weighting of the accepted material stream property, that is, a higher required purity, leads to higher values of the extended space parameter.
Therefore, objects are ejected more generously, with the risk of ejecting additional neighboring, actually accepting objects.
The reaction lines remained relatively constant.

\begin{table}[tbp]
\caption{Resulting process parameters $X$ after the optimization of the combined expected improvement (EI) for different weights $w_a$ (weight accept) and $w_r$ (weight reject).}
\begin{center}
\begin{tabular}{|c|c|c|c|c|}
\hline
\textbf{Weight}&\textbf{Weight}&\multicolumn{3}{|c|}{\textbf{$\arg\max_X\ \text{combined\ EI}(X)$}} \\
\cline{3-5} 
\textbf{accept} &\textbf{reject} & \textbf{\textit{Reaction Lines}}& \textbf{\textit{Ex. Time}}& \textbf{\textit{Ex. Space}} \\
\hline
1 & 0 & 15.67 & 1.37 & 2.29 \\
\hline
0 & 1 & 14.57 & 0.82 & 6.04 \\
\hline
0.7 & 0.3 & 14.90 & 2.99 & 4.21 \\
\hline
0.3 & 0.7 & 14.90 & 1.29& 7.92 \\
\hline
0.5 & 0.5 & 14.76 & 1.80 & 6.20 \\
\hline
\end{tabular}
\label{tab2}
\end{center}
\end{table}

The process parameters of an SBS system were successfully optimized based on a starting set of parameters and experiments using the selected sorting problem as an example.
In addition, the results of the different prioritization of the accepted or rejected streams are shown.
The resulting parameter values are reasonable, considering the optimization goal.

\section{Conclusion}
This study presents an optimization approach for the process parameters of an SBS system. The surrogate-based BO method uses GPR to model the sorting system process.
The proposed strategy allows the successful incorporation of uncertainties in measuring the sorting results and accuracy in calculating the models.
In addition, a method is presented in which the requirements can be placed on the purity of the resulting material streams after sorting.
Three exemplary process parameters were successfully optimized using an example sorting scenario, and the method's advantages were demonstrated and discussed.
Starting with a minimum initial parameter set and sorting experiments, an optimum parameter configuration was found in only a few steps.

The theoretical use of the optimization for an inline parameter adjustment based on, for example, a change in material flow was indicated. Practical testing and implementation using an example of a defined scenario offer opportunities for further research.

\bibliographystyle{IEEEtran}
\bibliography{references.bib}

\begin{thebibliography}{10}
\providecommand{\url}[1]{#1}
\csname url@samestyle\endcsname
\providecommand{\newblock}{\relax}
\providecommand{\bibinfo}[2]{#2}
\providecommand{\BIBentrySTDinterwordspacing}{\spaceskip=0pt\relax}
\providecommand{\BIBentryALTinterwordstretchfactor}{4}
\providecommand{\BIBentryALTinterwordspacing}{\spaceskip=\fontdimen2\font plus
\BIBentryALTinterwordstretchfactor\fontdimen3\font minus \fontdimen4\font\relax}
\providecommand{\BIBforeignlanguage}[2]{{%
\expandafter\ifx\csname l@#1\endcsname\relax
\typeout{** WARNING: IEEEtran.bst: No hyphenation pattern has been}%
\typeout{** loaded for the language `#1'. Using the pattern for}%
\typeout{** the default language instead.}%
\else
\language=\csname l@#1\endcsname
\fi
#2}}
\providecommand{\BIBdecl}{\relax}
\BIBdecl

\bibitem{maier2024survey}
G.~Maier, R.~Gruna, T.~Längle, and J.~Beyerer, ``A survey of the state of the art in sensor-based sorting technology and research,'' \emph{IEEE Access}, vol.~12, pp. 6473--6493, 2024.

\bibitem{gulcan2020novel}
E.~G{\"u}lcan, ``A novel approach for sensor based sorting performance determination,'' \emph{Minerals Engineering}, vol. 146, 2020.

\bibitem{forrester2008engineering}
A.~Forrester, A.~Sobester, and A.~Keane, \emph{Engineering design via surrogate modelling: a practical guide}.\hskip 1em plus 0.5em minus 0.4em\relax John Wiley \& Sons, 2008.

\bibitem{jones1998efficient}
D.~R. Jones, M.~Schonlau, and W.~J. Welch, ``Efficient global optimization of expensive black-box functions,'' \emph{Journal of Global optimization}, vol.~13, pp. 455--492, 1998.

\bibitem{snoek2012practical}
J.~Snoek, H.~Larochelle, and R.~P. Adams, ``Practical bayesian optimization of machine learning algorithms,'' \emph{Advances in neural information processing systems}, vol.~25, 2012.

\bibitem{li2017rapid}
C.~Li, D.~Rub{\'\i}n~de Celis~Leal, S.~Rana, S.~Gupta, A.~Sutti, S.~Greenhill, T.~Slezak, M.~Height, and S.~Venkatesh, ``Rapid bayesian optimisation for synthesis of short polymer fiber materials,'' \emph{Scientific reports}, vol.~7, no.~1, p. 5683, 2017.

\bibitem{nagai2022sample}
K.~Nagai, T.~Osa, G.~Inoue, T.~Tsujiguchi, T.~Araki, Y.~Kuroda, M.~Tomizawa, and K.~Nagato, ``Sample-efficient parameter exploration of the powder film drying process using experiment-based bayesian optimization,'' \emph{Scientific reports}, vol.~12, no.~1, p. 1615, 2022.

\bibitem{maier2020self}
M.~Maier, A.~Rupenyan, C.~Bobst, and K.~Wegener, ``Self-optimizing grinding machines using gaussian process models and constrained bayesian optimization,'' \emph{The International Journal of Advanced Manufacturing Technology}, vol. 108, pp. 539--552, 2020.

\bibitem{PFROMMER2018426}
J.~Pfrommer, C.~Zimmerling, J.~Liu, L.~Kärger, F.~Henning, and J.~Beyerer, ``Optimisation of manufacturing process parameters using deep neural networks as surrogate models,'' \emph{Procedia CIRP}, vol.~72, pp. 426--431, 2018, 51st CIRP Conference on Manufacturing Systems.

\bibitem{ZIMMERLING2022110423}
C.~Zimmerling, C.~Poppe, O.~Stein, and L.~Kärger, ``Optimisation of manufacturing process parameters for variable component geometries using reinforcement learning,'' \emph{Materials \& Design}, vol. 214, p. 110423, 2022.

\bibitem{pascoe_prediction_2015}
R.~D. Pascoe, R.~Fitzpatrick, and J.~R. Garratt, ``Prediction of automated sorter performance utilising a {Monte} {Carlo} simulation of feed characteristics,'' \emph{Minerals Engineering}, vol.~72, pp. 101--107, Mar. 2015.

\bibitem{bauer2022towards}
A.~Bauer, G.~Maier, M.~Reith-Braun, H.~Kruggel-Emden, F.~Pfaff, R.~Gruna, U.~Hanebeck, and T.~Längle, ``Towards a feed material adaptive optical belt sorter: A simulation study utilizing a {DEM}--{CFD} approach,'' \emph{Powder Technology}, vol. 411, p. 117917, 2022.

\bibitem{vieth2023improving}
J.~Vieth, M.~Reith-Braun, A.~Bauer, F.~Pfaff, G.~Maier, R.~Gruna, T.~Längle, H.~Kruggel-Emden, and U.~D. Hanebeck, ``Improving accuracy of optical sorters using closed-loop control of material recirculation,'' in \emph{2023 American Control Conference (ACC)}, 2023, pp. 3257--3263.

\bibitem{walker2025stochastic}
M.~Walker, M.~Reith-Braun, A.~Bauer, F.~Pfaff, G.~Maier, R.~Gruna, T.~Längle, J.~Beyerer, H.~Kruggel-Emden, and U.~D. Hanebeck, ``Stochastic optimal control of an optical sorter with material recirculation,'' \emph{IEEE Transactions on Control Systems Technology}, vol.~33, no.~1, pp. 354--368, 2025.

\bibitem{kronenwett2024sensor}
F.~Kronenwett, G.~Maier, N.~Leiss, R.~Gruna, V.~Thome, and T.~L{\"a}ngle, ``Sensor-based characterization of construction and demolition waste at high occupancy densities using synthetic training data and deep learning,'' \emph{Waste Management \& Research}, vol.~42, no.~9, pp. 788--796, 2024.

\bibitem{shahriari2015taking}
B.~Shahriari, K.~Swersky, Z.~Wang, R.~P. Adams, and N.~De~Freitas, ``Taking the human out of the loop: A review of bayesian optimization,'' \emph{Proceedings of the IEEE}, vol. 104, no.~1, pp. 148--175, 2015.

\bibitem{10.7551/mitpress/3206.001.0001}
C.~E. Rasmussen and C.~K.~I. Williams, \emph{Gaussian Processes for Machine Learning}.\hskip 1em plus 0.5em minus 0.4em\relax The MIT Press, 11 2005.

\bibitem{brochu2010tutorial}
E.~Brochu, V.~M. Cora, and N.~De~Freitas, ``A tutorial on bayesian optimization of expensive cost functions, with application to active user modeling and hierarchical reinforcement learning,'' \emph{arXiv preprint arXiv:1012.2599}, 2010.

\end{thebibliography}

\end{document}